\documentclass[11pt,letterpaper]{article}
\usepackage{emnlp2017}
\usepackage{times}
\usepackage{latexsym}
\usepackage{arydshln}
\usepackage{amsmath}
\usepackage{url}
\usepackage{graphicx}
\usepackage{adjustbox}
\usepackage{caption}
\usepackage{subcaption}
\usepackage{amssymb}
\usepackage{multirow}
\usepackage{rotating}
\newcommand*\rot{\rotatebox{90}}
\newcommand\Tstrut{\rule{0pt}{2.3ex}}       
\newcommand\Bstrut{\rule[-0.9ex]{0pt}{0pt}} 

\emnlpfinalcopy

\def\emnlppaperid{387}

\usepackage{boxedminipage}
\usepackage{multirow}

\newcommand{\bleu}{\textsc{\small bleu}}
\newcommand{\rouge}{\textsc{\small rouge}}
\newcommand{\meteor}{\textsc{\small meteor}}
\newcommand{\cider}{\textsc{\small cide}r}
\newcommand{\ter}{\textsc{\small ter}}
\newcommand{\nist}{\textsc{\small nist}}
\newcommand{\lepor}{\textsc{\small lepor}}
\newcommand{\fres}{\textsc{\small re}}
\newcommand{\simil}{\textsc{\small sim}}

\newcommand{\vlachos}{\textsc{lols}}
\newcommand{\dusek}{\textsc{\small TGen}}
\newcommand{\wen}{\textsc{rnnlg}}

\newcommand{\sfh}{\textsc{\small SFHotel}}
\newcommand{\sfr}{\textsc{\small SFRest}}
\newcommand{\bagel}{\textsc{\small Bagel}}

\newcommand{\p}{\phantom{-}} 

\newcommand{\ignore}[1]{}

\usepackage{xcolor}
\usepackage[normalem]{ulem}
\definecolor{darkgreen}{rgb}{0.0, 0.5, 0.0}
\def\OD#1{{\color{darkgreen}OD: \it #1}}
\def\ODdel#1{\bgroup\markoverwith{\textcolor{darkgreen}{\rule[0.5ex]{2pt}{1pt}}}\ULon{#1}}

\def\VRdel#1{\bgroup\markoverwith{\textcolor{red}{\rule[0.5ex]{2pt}{1pt}}}\ULon{#1}}

\title{The BLEU Screen of Death: Why We Need New Metrics for NLG}
\title{Death by BLEU: Why We Need New Metrics for NLG}
\title{From {\textsc{bleu}} to {\textsc{rainbow}}: Why We Need New Metrics for NLG.}
\title{Why We Need New Evaluation Metrics for NLG}

\author{Jekaterina Novikova, \ Ond\v{r}ej Du{\v{s}}ek, \ Amanda Cercas Curry \and Verena Rieser \\
  School of Mathematical and Computer Sciences \\
  Heriot-Watt University, Edinburgh \\
  {\tt j.novikova, o.dusek, ac293, v.t.rieser@hw.ac.uk} 
  }
  
\date{}

\begin{document}

\maketitle

\begin{abstract}
The majority of NLG evaluation relies on automatic metrics, such as \bleu. In this paper, we motivate the need for novel, system- and data-independent automatic evaluation methods:
We investigate a wide range of metrics, including state-of-the-art word-based and novel grammar-based ones, 
and  demonstrate that they only weakly reflect human judgements of system outputs as generated by data-driven, end-to-end NLG.
We also show that metric performance is data- and system-specific. 
Nevertheless, our results also suggest that 
automatic metrics perform reliably at system-level and can support system development by finding cases where a system  performs poorly.%

\end{abstract}

\section{Introduction} 
Automatic evaluation measures, such as \bleu{} \cite{papineni2002bleu}, are used with increasing frequency to evaluate Natural Language Generation (NLG) systems: Up to 60\% of  NLG research published between 2012--2015 relies on automatic metrics \cite{gkatzia:enlg2015}.
Automatic evaluation is popular because it is cheaper and faster to run than human evaluation, and it is needed for automatic benchmarking and tuning of algorithms. 
The use of such metrics is, however, only sensible if they are known to be sufficiently correlated with human preferences. 
This is rarely the case, as shown by various studies in NLG  (\citealp{stent2005evaluating}; \citealp{Belz:EACL06,belz:CL2009}), as well as in  related fields, such as  dialogue systems \cite{Liu:EMNLP2016}, machine translation (MT) \cite{callison:eacl2006}, and image captioning \cite{keller:acl2014,erdem:eacl2017}. 
This paper follows on from the above previous work and presents another evaluation study into automatic metrics with the aim to firmly establish the need for new metrics. 
We consider this paper to be the most complete study to date, across metrics, systems, datasets and domains, focusing on recent advances in data-driven NLG.
 In contrast to previous work, we are the first to:

\noindent $\bullet$  Target end-to-end data-driven NLG,  
where we compare 3 different approaches. 
In contrast to NLG methods evaluated in previous work, our  
systems can produce ungrammatical output by (a) generating word-by-word, and (b) learning from noisy data.

\noindent $\bullet$ Compare a large number of 21
automated metrics, including novel grammar-based ones. 

\noindent $\bullet$ Report results on two different domains and three different datasets, which allows us to draw more general conclusions.

\noindent $\bullet$ Conduct a detailed error analysis, which 
suggests that, while metrics can be  reasonable indicators at the system-level, they are not reliable at the sentence-level.

\noindent $\bullet$ Make all associated code and data publicly available, including detailed analysis results.\footnote{Available for download at: \url{https://github.com/jeknov/EMNLP_17_submission}}

\section{End-to-End NLG Systems}\label{sec:systems} 

In this paper, we focus on recent end-to-end, data-driven NLG methods, which jointly learn sentence planning and surface realisation from non-aligned data (\citealp{jurcicek:2015:ACL,wen:emnlp2015,Mei:NAACL2015,Wen:NAACL16}; \citeauthor{SharmaHSSB16}, \citeyear{SharmaHSSB16}; \citealp{Dusek:ACL16}, \citealp{vlachos:coling2016}).
These approaches do not require costly semantic alignment between Meaning Representations (MR) and human references (also referred to as ``ground truth" or ``targets"), but are based on parallel datasets, which can be collected in sufficient quality and quantity using effective crowdsourcing techniques, e.g.\ \cite{novikova:INLG2016}, and as such, enable rapid development of NLG components in new domains. 
In particular, we compare the performance of the following systems:

\noindent $\bullet$ {\bf \wen :}{}\footnote{\url{https://github.com/shawnwun/RNNLG}}
The system by \citet{wen:emnlp2015} uses a Long Short-term Memory (LSTM) network to 
jointly address sentence planning and surface realisation. 
It augments each LSTM cell with a gate that conditions it on the input MR, which allows it to keep track of MR contents generated so far.

\noindent $\bullet$ {\bf \dusek :}\footnote{\url{https://github.com/UFAL-DSG/tgen}}
The system by \citet{jurcicek:2015:ACL} learns to incrementally generate deep-syntax dependency trees of candidate sentence plans (i.e.\ which MR elements to mention and the overall sentence structure). Surface realisation is performed using a separate, domain-independent rule-based module.

\noindent $\bullet$ {\bf \vlachos :}\footnote{\url{https://github.com/glampouras/JLOLS_NLG}}
The system by \citet{vlachos:coling2016} learns sentence planning and surface realisation using Locally Optimal Learning to Search (\vlachos), an imitation learning framework which learns using \bleu{} and \rouge{} as non-decomposable loss functions. 


\begin{table}[tb]
\begin{center}
\footnotesize
\begin{tabular}{c|r|r|r|r}
\multirow{2}{*}{\bf System} & \multicolumn{3}{c|}{\bf Dataset} & \multirow{2}{*}{Total} \\
 & \bagel & \sfr & \sfh & \Bstrut \\\hline\hline
\vlachos & 202   & 581    & 398   & 1,181\Tstrut \\
\wen     & -     & 600    & 477   & 1,077 \\
\dusek   & 202   & -      & -     &   202\Bstrut \\\hdashline[0.5pt/2pt]
Total    & 404   & 1,181  & 875   & 2,460\Tstrut \\
\end{tabular}
\end{center}
\caption{Number of NLG system outputs from different datasets and systems used in this study.}\label{tab:overview}
\vspace{-0.5cm}
\end{table}

\section{Datasets}\label{sec:data} 

We consider the following crowdsourced datasets, which target utterance generation for spoken dialogue systems. 
Table \ref{tab:overview} shows the number of system outputs for each dataset.
Each data instance consists of one MR and one 
or more natural language references as produced by humans, such as the following example, taken from the \bagel{} dataset:%
\footnote{Note that we use lexicalised versions of \sfh{} and \sfr{} and a partially lexicalised version of \bagel, where proper names and place names are replaced by placeholders (``X''), in correspondence with the outputs generated by the systems, as provided by the system authors.}
\vspace{0.2cm}

\noindent\begin{boxedminipage}{\columnwidth}
 \small {\bf MR:} inform(name=X, area=X, pricerange=moderate, type=restaurant)\\
 {\bf Reference:} ``{\em X is a moderately priced restaurant in X.}''
 \end{boxedminipage}   
\vspace{0.1cm}
 
\noindent  $\bullet$ {\bf \sfh{} \& \sfr}  \cite{wen:emnlp2015} provide information about hotels and restaurants in San Francisco. There are 8 system dialogue act types, such as \emph{inform}, 
\emph{confirm}, 
\emph{goodbye}
 etc. 
 Each domain contains 12 attributes, where some are common to both domains,
 such as \emph{name, type, pricerange, address, area,} etc., 
 and the others are domain-specific, e.g.\ \emph{food} and \emph{kids-allowed} for restaurants; 
 \emph{hasinternet} and \emph{dogs-allowed} for hotels.
For each domain, around 5K human references were collected 
with 2.3K unique human utterances for \sfh{} and 1.6K for \sfr. The number of unique system outputs produced is 1181 for \sfr{} and 875 for \sfh{}. 

\noindent $\bullet$ {\bf \bagel} \cite{mairesse:acl2010} provides information about restaurants in Cambridge. The dataset contains 202 aligned pairs of MRs and 2 corresponding references each. 
The domain is a subset of \sfr, including only the  {\em inform} act and 8 attributes.

\section{Metrics}\label{sec:metrics} 

\subsection{Word-based Metrics (WBMs)}\label{sec:wbm}

NLG evaluation has borrowed a number of automatic metrics from related fields, such as MT, summarisation or image captioning, 
which compare
output texts generated by systems to ground-truth references produced by humans. We refer to this group as  
 word-based metrics. 
 In general, the higher these scores are, the better or more similar to the human references the output is.\footnote{Except for \ter{} whose scale is reversed.}
 The following order reflects the degree these metrics move from simple $n$-gram overlap to also considering term frequency (TF-IDF) weighting and semantically similar words.

\noindent $\bullet$ {\bf Word-overlap Metrics (WOMs):} 
We consider frequently used metrics, including 
 {\bf \ter} \cite{ter},
{\bf \bleu} \cite{papineni2002bleu},
{\bf \rouge} \cite{lin2004rouge},
{\bf \nist} \cite{nist},
{\bf \lepor} \cite{lepor},
{\bf \cider} \cite{cider}, and
{\bf \meteor} \cite{meteor}.

\noindent $\bullet$ {\bf Semantic Similarity (\simil)}
: We calculate the Semantic Text Similarity measure designed by \citet{han2013umbc}. 
This measure is based on distributional similarity and Latent Semantic Analysis (LSA) and is further complemented with semantic relations extracted from WordNet. 
\subsection{Grammar-based metrics (GBMs)}
\label{sec:gbm}
Grammar-based measures have been explored in related fields, such as MT  \cite{Gimenez:2008} or grammatical error correction \cite{napoles2016}, and, in contrast to WBMs, do not rely on ground-truth references. 
To our knowledge, we are the first to consider 
 GBMs for sentence-level NLG evaluation. We focus on two important properties of texts here -- readability and grammaticality:
 
\noindent $\bullet$ {\bf Readability} 
quantifies the difficulty with which a reader understands a text, as used for e.g.\ evaluating summarisation \cite{kan2001applying} or text simplification \cite{readabilitySI:2014}. 
We measure readability by the 
Flesch Reading Ease score ({\bf \fres})~\cite{flesch1979write},
which  calculates 
a ratio between the number of characters per sentence, the number of words per sentence,
and 
the number of syllables per word. Higher \fres{} score indicates a less complex utterance that is easier to read and understand.
We also consider related measures, such as 
 characters per utterance ({\bf len}) and per word ({\bf cpw}), words per sentence ({\bf wps}), syllables per sentence ({\bf sps}) and per word ({\bf spw}), as well as polysyllabic words per utterance ({\bf pol}) and per word ({\bf ppw}). The higher these scores, the more complex the utterance. 

\noindent $\bullet$ {\bf Grammaticality:} 
In contrast to previous NLG methods, our corpus-based end-to-end systems can produce ungrammatical output by (a) generating word-by-word, and (b) learning from noisy data.
As a first approximation of
grammaticality, we measure the number of misspellings (\textbf{msp}) and the parsing score as returned by the Stanford parser (\textbf{prs}). The lower the \textbf{msp}, the more grammatically correct an utterance is. 
The  
Stanford parser score is not designed to measure grammaticality, however, it will generally prefer a grammatical parse to a non-grammatical one.\footnote{\url{http://nlp.stanford.edu/software/parser-faq.shtml}} Thus, lower parser scores indicate less grammatically-correct utterances. 
In future work, we aim to use specifically designed grammar-scoring functions, e.g.\ \cite{napoles2016}, once they become publicly available.

\section{Human Data Collection}\label{sec:datacollection} 

To collect human rankings, 
we presented the MR together with 2 utterances generated by different systems side-by-side to crowdworkers, which were asked to score each utterance on a 6-point Likert scale for:

\noindent $\bullet$ {\bf Informativeness:} {\em Does the utterance  provide all the useful information from the meaning representation?} 

\noindent$\bullet$ {\bf Naturalness:} {\em Could the utterance have been produced by a native speaker?} 

\noindent$\bullet$ {\bf Quality:} {\em How do you judge the overall quality of the utterance in terms of its grammatical correctness and fluency?} 

Each system output (see~Table~\ref{tab:overview}) was scored by 3 different crowdworkers.
To reduce participants' bias, the order of appearance of utterances produced by each system was randomised and crowdworkers were restricted to evaluate a maximum of 20 utterances. 
The crowdworkers were selected from  English-speaking countries only, based on their IP addresses, and asked to confirm 
that English was their native language.

To assess the reliability of ratings, we calculated the intra-class correlation coefficient (ICC), which measures inter-observer reliability on ordinal data for more than two raters \cite{landis1977measurement}. The overall ICC across all three datasets is 0.45 ($p<0.001$), which corresponds to a moderate agreement. 
In general, we find consistent differences in inter-annotator agreement per system and dataset, 
with lower agreements for \vlachos{} than for \wen{} and \dusek. Agreement is highest for the \sfh{} dataset, followed by \sfr{} and \bagel{} (details provided in supplementary material). 

\section{System Evaluation}\label{sec:evaluation} 

\begin{table*}[h!]
\centering
\begin{adjustbox}{max width=1\textwidth}
\begin{tabular}{l|cc|cc|cc}
 & \multicolumn{2}{c|}{\bagel} & \multicolumn{2}{c|}{\sfh} & \multicolumn{2}{c}{\sfr} \\
\bf metric & \multicolumn{1}{c}{\dusek} & \multicolumn{1}{c|}{\vlachos} & \multicolumn{1}{c}{\wen} & \multicolumn{1}{c|}{\vlachos} & \multicolumn{1}{c}{\wen} & \multicolumn{1}{c}{\vlachos} \\
\hline \hline
WOMs &  & More overlap & More overlap* &  & More overlap* & \Tstrut \\
\simil & More similar &  & More similar* &  &  & More similar \\
\hdashline[1pt/3pt]
GBMs & Better grammar(*) &  & Better grammar(*) &  & Better grammar &  \\
\fres &  & More complex* &  & More complex* &  & More complex*\Bstrut\\
\hline
inform & 4.77(Sd=1.09) & \textbf{4.91}(Sd=1.23) & \textbf{5.47*}(Sd=0.81) & 5.27(Sd=1.02) & \textbf{5.29*}(Sd=0.94) & 5.16(Sd=1.07)\Tstrut\\
natural & \textbf{4.76}(Sd=1.26) & 4.67(Sd=1.25) & \textbf{4.99*}(Sd=1.13) & 4.62(Sd=1.28) & \textbf{4.86}\phantom{*}(Sd=1.13) & 4.74(Sd=1.23) \\
quality & \textbf{4.77}(Sd=1.19) & 4.54(Sd=1.28) & \textbf{4.54}\phantom{*}(Sd=1.18) & 4.53(Sd=1.26) & 4.51\phantom{*}(Sd=1.14) & \textbf{4.58}(Sd=1.33)
\end{tabular}
\end{adjustbox}
\caption{System performance per dataset (summarised over metrics), where ``*'' denotes $p<0.05$ for all the metrics and ``(*)'' shows significance on $p<0.05$ level for the majority of the metrics.}
\label{tab:sys-eval}
\end{table*}

Table \ref{tab:sys-eval} summarises the
 individual systems' overall corpus-level performance in terms of automatic and human scores (details are provided in the supplementary material). 

All WOMs 
produce similar results, with \simil{}
showing different results for the restaurant domain (\bagel{} and \sfr).
Most GBMs show the same trend (with different levels of statistical significance), but \fres{} is showing inverse results.
 System performance is dataset-specific: 
For WBMs,
 the \vlachos{} system consistently produces better results on \bagel{} compared to \dusek, 
 while for \sfr{} and \sfh, \vlachos{} is outperformed by \wen{} in terms of WBMs.
We observe that human {\em informativeness} ratings follow the same pattern as WBMs, while the average similarity score (\simil) seems to be related to human {\em quality} ratings.

Looking at GBMs, we observe that they seem to be related to {\em naturalness} and {\em quality} ratings.
Less complex utterances, as measured by readability (\fres) and word length (cpw), have higher {\em naturalness} ratings. 
More complex utterances, as measured in terms of their length (len), number of words (wps), syllables (sps, spw) and polysyllables (pol, ppw), have lower {\em quality} evaluation. 
 Utterances measured as more grammatical 
 are on average evaluated higher in terms of {\em naturalness}.

These initial results suggest a relation between automatic metrics and human ratings at system level. However, average scores can be misleading, as they do not identify worst-case scenarios.
This leads us to inspect the correlation of human and automatic metrics for each MR-system output pair at utterance level.

\begin{table*}[ht]
\centering
\begin{adjustbox}{max width=1\textwidth}
\begin{tabular}{cl|ll|ll|ll}
 &  & \multicolumn{2}{c|}{\bagel} & \multicolumn{2}{c|}{\sfh} & \multicolumn{2}{c}{\sfr} \\
 &  & \dusek & \vlachos & \wen & \vlachos & \wen & \vlachos \\
\hline \hline
Best & inform. & \p0.30* (\bleu{\small-1}) & \p0.20* (\rouge) & \p0.09 (\bleu{\small-1}) & \p0.14* (\lepor) & \p0.13* (\simil) & \p0.28* (\lepor) \\
WBM & natural. & -0.19* (\ter) & -0.19* (\ter) & \p0.10* (\meteor) & -0.20* (\ter) & \p0.17* (\rouge) & \p0.19* (\meteor) \\
 & quality & -0.16* (\ter) & \p0.16* (\meteor) & \p0.10* (\meteor) & -0.12* (\ter) & \p0.09* (\meteor) & \p0.18* (\lepor) \\
 \hline
Best & inform. & \p0.33* (wps) & \p0.16* (ppw) & -0.09 (ppw) & \p0.13* (cpw) & \p0.11* (len) & \p0.21* (len) \\
GBM & natural. & -0.25* (len) & -0.28* (wps) & -0.17* (len) & -0.18* (sps) & -0.19* (wps) & -0.21* (sps) \\
 & quality & -0.19* (cpw) & \p0.31* (prs) & -0.16* (ppw) & -0.17* (spw) & \p0.11* (prs) & -0.16* (sps) \\
\end{tabular}
\end{adjustbox}
\caption{Highest absolute Spearman correlation between metrics and human ratings,  
 with ``*'' denoting $p<0.05$ (metric with the highest absolute value of $\rho$ given in brackets).}
\label{tab:corr_rf}
\end{table*}

\begin{figure*}[h!]
\centering
\vspace{-2mm}
\includegraphics[width=0.99\textwidth]{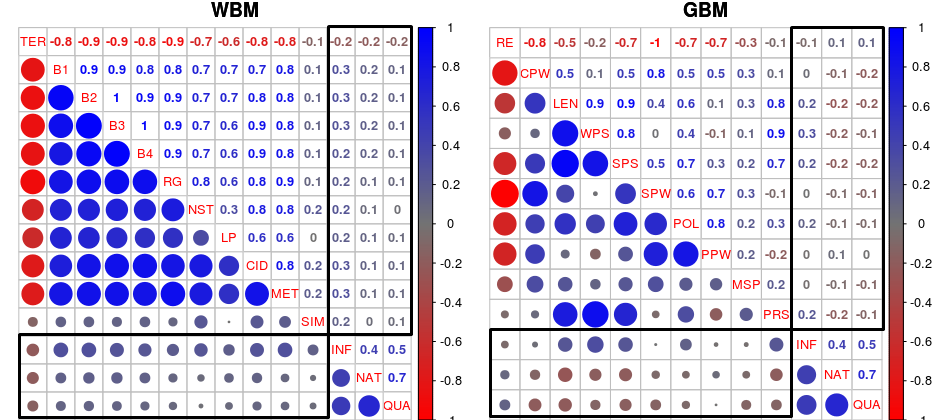}
\vspace{2mm}
\caption{Spearman correlation results for \dusek{} on \bagel. 
Bordered area shows correlations between human ratings and automatic metrics, the rest shows correlations among the metrics. Blue colour of circles indicates positive correlation, while red indicates negative correlation. The size of circles denotes the correlation strength.} 
\label{fig:corr}
\end{figure*}

\begin{figure}[h]
\centering
\includegraphics[width=0.49\textwidth]{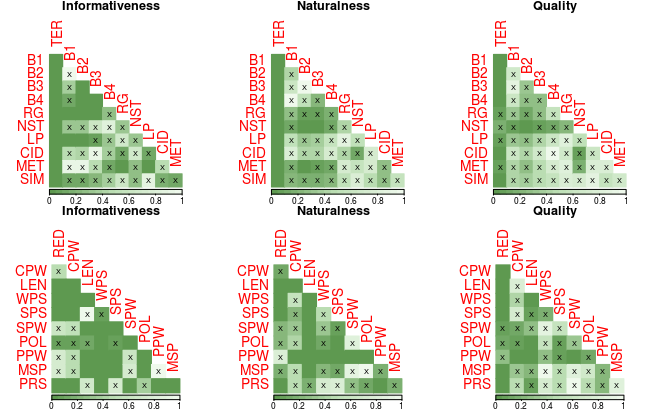}
\caption{Williams test results: X represents a {\em non}-significant difference between correlations ($p<0.05$; top: WBMs, bottom: GBMs).} 
\label{fig:wil}
\end{figure}

\section{Relation of Human and Automatic Metrics}\label{sec:correlation}

\subsection{Human Correlation Analysis}\label{ssec:correlation} 

We 
calculate the correlation between automatic metrics and human ratings
using the Spearman coefficient ($\rho$). 
We split the data per dataset and system in order to make valid pairwise comparisons. 
To handle outliers within human ratings, we use the median score of the three human raters.%
\footnote{As an alternative to using the median human judgment for each item,
a more effective way to use all the human judgments could be to use
\citet{hovy_learning_2013}'s MACE tool for inferring the reliability of
judges.}
Following \citet{erdem:eacl2017}, we use the Williams' test \cite{williamstest} to determine significant differences between correlations. 
%
Table~\ref{tab:corr_rf} summarises the utterance-level correlation results between automatic metrics and human ratings, listing the best (i.e.\ highest absolute $\rho$) results for each type of metric (details provided in supplementary material). 
 Our results  
suggest that:

\noindent $\bullet$ 
In sum, no metric produces 
an even moderate correlation with human ratings, 
 independently of dataset, system, or aspect of human rating. This contrasts with our initially promising results on the system level (see Section \ref{sec:evaluation}) 
 and will be further discussed in Section \ref{sec:error}. Note that similar inconsistencies between document- and sentence-level evaluation results are observed in MT \cite{specia:MT2010}. 

\noindent $\bullet$ Similar to our 
results in Section \ref{sec:evaluation}, we find that WBMs show better correlations to human ratings of {\em informativeness} 
(which reflects content selection), 
whereas GBMs show better correlations to {\em quality} and {\em naturalness}. 

\noindent $\bullet$ Human ratings for {\em informativeness}, {\em naturalness} and {\em quality} are highly correlated with each other, with the highest correlation between the latter two ($\rho=0.81$) reflecting that they both target surface realisation.

\noindent $\bullet$ All WBMs produce similar results (see Figure~\ref{fig:corr} and \ref{fig:wil}): They are strongly correlated with each other, and most of them produce correlations with human ratings which are {\em not} significantly different from each other. 
GBMs, on the other hand, show greater diversity. 

\noindent $\bullet$ Correlation results are system- and dataset-specific (details provided in supplementary material). 
We observe the  highest correlation for \dusek{} on \bagel{} (Figures~\ref{fig:corr} and \ref{fig:wil}) and \vlachos{} on \sfr, whereas \wen{} often shows low 
correlation between metrics and human ratings. 
This lets us conclude that WBMs and GBMs are sensitive to different systems and datasets.

 \noindent $\bullet$ The highest positive correlation 
  is observed between the number of words (wps) and {\em informativeness} for the \dusek{} system on \bagel{} ($\rho=0.33$,\linebreak $p<0.01$, see Figure~\ref{fig:corr}){. 
However, the wps metric (amongst most others) is not robust across systems and datasets: Its correlation on other datasets is very weak, ($\rho\le .18$) and its correlation with informativeness ratings of \vlachos{} outputs is insignificant.

\noindent $\bullet$ As a sanity check, we also measure a random score $[0.0,1.0]$ which proves to have a close-to-zero correlation with human ratings (highest $\rho=0.09$).

\subsection{Accuracy of Relative Rankings}\label{ssec:ranking} 

We now evaluate a more coarse measure, namely the metrics' ability to predict relative human ratings. That is, we compute the score of each metric for
two system output sentences corresponding to the same MR. 
The prediction of a metric is correct if it orders the sentences in the same way as median human ratings (note that ties are allowed).
Following previous work \cite{cider,erdem:eacl2017}, we mainly concentrate on WBMs. 
 Results summarised in Table \ref{tab:accuracy} 
 show that most metrics' performance is not significantly different from that of a random score 
 (Wilcoxon signed rank test). 
 While the random score fluctuates between 25.4--44.5\% prediction accuracy, the metrics achieve an accuracy of between 30.6--49.8\%.
 Again, the performance of the metrics is dataset-specific:
 Metrics perform best on \bagel{} data; for \sfh, metrics show mixed performance while for \sfr, metrics perform worst.
 
 \ignore{
 The metrics perform best on \bagel{} data, with all metrics 
 performing significantly better than random.
 For \sfh, metrics show mixed performance: 
For informativeness, all but \nist{} are better than the random metric, and for naturalness only \meteor{} outperforms random.
None of the metrics can outperform random for quality (with \simil{} and \fres{} being significantly worse than random).
For \sfr, metrics perform worst, with only \simil{} for informativeness and \lepor{} for naturalness outperforming random.
\OD{I'd shorten the talk about datasets and leave it to the table... just mention that the performance is dataset-specific?}
}

\begin{table}[h!]
\centering
\begin{adjustbox}{max width=0.5\textwidth}
\begin{tabular}{ll|l|l|l}
\multicolumn{2}{l|}{} & \bf informat. & \bf naturalness & \bf quality \\
\hline
\hline
\bagel & \begin{tabular}[c]{@{}l@{}}raw \\ data\end{tabular} & \begin{tabular}[c]{@{}l@{}}\ter, \bleu{\small 1-4}, \\ \rouge, \nist, \\ \lepor, \cider, \\ \meteor, \simil \end{tabular} & \begin{tabular}[c]{@{}l@{}}\ter, \bleu{\small 1-4}, \\ \rouge, \nist, \\ \lepor, \cider, \\ \meteor, \simil \end{tabular} & \begin{tabular}[c]{@{}l@{}}\ter, \bleu{\small 1-4}, \\ \rouge, \nist, \\ \lepor, \cider, \\ \meteor, \simil \end{tabular} \\
\hline
\sfh & \begin{tabular}[c]{@{}l@{}}raw \\ data\end{tabular} & \begin{tabular}[c]{@{}l@{}}\ter, \bleu{\small 1-4}, \\ \rouge, \lepor, \\ \cider, \meteor, \\  \simil \end{tabular} & \meteor & N/A \\
\hline
\multirow{2}{*}{\sfr} & \begin{tabular}[c]{@{}l@{}}raw \\ data\end{tabular} &  \simil & \lepor & N/A \\
\cdashline{2-5}[1pt/3pt]
 & \begin{tabular}[c]{@{}l@{}}quant. \\ data\end{tabular} & \begin{tabular}[c]{@{}l@{}}\ter, \bleu{\small 1-4}, \\ \rouge, \nist, \\ \lepor, \cider, \\ \meteor \\ \simil \end{tabular} & N/A & N/A
\end{tabular}
\end{adjustbox}
\caption{Metrics predicting relative human rating with significantly higher accuracy than a random baseline.}
\label{tab:accuracy}
\end{table}

\noindent {\bf Discussion:}
 Our data differs from the one used in previous work \cite{cider,erdem:eacl2017}, which uses 
explicit relative rankings 
(``{\em Which output do you prefer?}"), whereas we compare two Likert-scale ratings. As such, we have 3 possible outcomes (allowing ties). 
This way, we can account for equally valid system outputs, which is one of the main drawbacks of forced-choice approaches \cite{hodosh2016focused}.
Our results are akin to previous work: 
\citet{erdem:eacl2017} report results between 60-74\% accuracy for binary classification on machine-machine data, which is comparable to our results for 3-way classification.

Still, we observe a mismatch between the ordinal human ratings and the continuous metrics. For example, humans might rate system A and system B both as a 6, whereas \bleu, for example, might assign 0.98 and 1.0 respectively, meaning that \bleu{} will declare system B as the winner. 
In order to account for this mismatch, we quantise our metric data to the same scale as the median scores from our human ratings.\footnote{Note that this mismatch can also be accounted for by continuous rating scales, as suggested by \citet{Belz:ACL2011}.} Applied to \sfr, where we previously got our worst results, we can see an improvement for predicting {\em informativeness}, where all WBMs now perform significantly better than the random baseline (see Table~\ref{tab:accuracy}).
In the future, we will investigate related discriminative approaches, e.g.\  \cite{hodosh2016focused,vinyals:gans}, where the task is simplified to distinguishing correct from incorrect output.
 
\section{Error Analysis}\label{sec:error} 

In this section, we attempt to uncover why automatic metrics perform so poorly.

\subsection{Scales} 

We first explore the hypothesis that metrics are good in distinguishing extreme cases, i.e.\ system outputs which are rated as clearly good or bad by the human judges, but do not perform well for utterances rated in the middle of the Likert scale, as suggested by \citet{erdem:eacl2017}.  
 We  `bin' our data into three groups: {\em bad}, which comprises low ratings ($\leq$2); {\em good}, comprising  high ratings ($\geq$5); and finally a group comprising {\em average} ratings. 
 
We find that utterances with low human ratings of {\em informativeness} and {\em naturalness} correlate significantly better ($p<0.05$) with automatic metrics than those with average and good human ratings. For example,  as shown in Figure~\ref{fig:corr-bins}, the correlation between WBMs and human ratings for utterances with low {\em informativeness} scores ranges between $0.3 \le \rho \le 0.5$ (moderate correlation), while the highest correlation for utterances of average and high informativeness barely reaches $\rho \le 0.2$ (very weak correlation).
The same pattern can be observed for correlations with {\em quality} and {\em naturalness} ratings.  

This discrepancy in correlation results between low and other user ratings, together with the fact that the majority of system outputs are rated 
``good" for informativeness (79\%), naturalness (64\%) and quality (58\%), whereas low ratings do not exceed 7\% in total, 
could explain why the overall correlations are low (Section \ref{sec:correlation})
despite the observed trends in relationship between average system-level performance scores (Section \ref{sec:evaluation}). 
It also explains why the \wen{} system, which 
contains very few instances of low user ratings, shows poor correlation between human ratings and automatic metrics. 

\begin{figure}[h]
\centering
\includegraphics[width=0.3\textwidth]{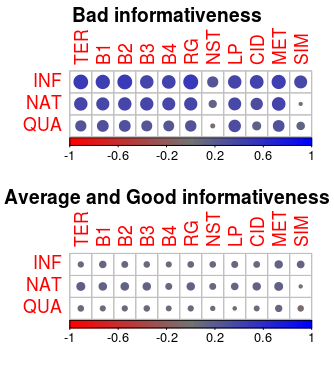}
\caption{Correlation between automatic metrics (WBMs) and human ratings for utterances of bad informativeness (top), and average and good informativeness (bottom).  } 
\label{fig:corr-bins}
\vspace{-0.5cm}
\end{figure}

 \begin{table*}[htp]
\small
\centering
\begin{adjustbox}{max width=1\textwidth}
\begin{tabular}{c|c|p{3.7cm}|p{3.5cm}|p{3.7cm}|c|c|c }
\hspace{-1mm}\bf No.\hspace{-1mm} & \bf system & \multicolumn{1}{c|}{\bf MR} &  \multicolumn{1}{c|}{\bf system output} &  \multicolumn{1}{c|}{\bf human reference} & \bf WOMs   & \bf \simil & \hspace{-1mm}\bf humans\Tstrut\hspace{-1mm} \\
\hline \hline
1& \vlachos & \raggedright\hangindent=0.5em\hangafter=1 inform(name = the
donatello, hasinternet = yes) & well
there is a hotel with internet access called the donatello &  the donatello has internet & 1.4	&5  & 6\Tstrut \\
\hline
2& \vlachos & \raggedright\hangindent=0.5em\hangafter=1 inform\_nomatch(area = embarcadero, kidsallowed\hspace{-0.2em}~=\hspace{-0.2em} yes, pricerange\hspace{-0.1em}~=\hspace{-0.1em}~expensive)\hspace{-0.2em} & i but i but i but i
but i but i but i but i but i but i & unfortunately i could not find any expensive restaurants in embarcadero that allow kids. & 1.1 & 1& 1\Tstrut \\
\hline
3& \dusek{} & \raggedright\hangindent=0.5em\hangafter=1 inform(name = X, area
= riverside, eattype = restaurant, food = fastfood,
pricerange = cheap) & x is a restaurant on the
riverside called located at the riverside and at is   & x is a cheap fastfood
restaurant located near the riverside & 2.4 & 4& 1 \Tstrut\\
\hline
4& \wen{}  & \raggedright\hangindent=0.5em\hangafter=1 inform\_nomatch(kidsallowed = yes, food = moroccan) & i am sorry, i did
not find any restaurants that allows kids and serve
moroccan. & sorry,
there are no restaurants allowing kids and serving
moroccan food & 1.85  & 4 & 5 \Tstrut\\
\end{tabular}
\end{adjustbox}
\vspace{-0.1cm}
\caption
{Example pairs of MRs and system outputs from our data, contrasting the average of word-overlap metrics (normalised in the 1-6 range) and semantic similarity (\simil) with human ratings (median of all measures).}
\label{tab:examples}
\vspace{-0.2cm}
\end{table*}%

\subsection{Impact of Target Data}
\label{ssec:impactData}

\noindent {\bf Characteristics of Data: }
In Section \ref{ssec:correlation}, we observed that datasets have a significant impact on how well automatic metrics reflect human ratings. A closer inspection shows that \bagel{} data differs significantly from \sfr{} and \sfh, both in terms of grammatical and MR properties.
\bagel{} has significantly shorter references both in terms of number of characters and words compared to the other two datasets.
Although being shorter, the words in \bagel{} references are significantly more often polysyllabic.
Furthermore,
\bagel{} only consists of utterances generated from \emph{inform} MRs, while \sfr{} and \sfh{} also have less complex MR types, such as {\em confirm}, {\em goodbye}, etc.
Utterances produced from \emph{inform} MRs are significantly longer and have a significantly higher correlation with human ratings of \emph{informativeness} and \emph{naturalness} than \emph{non-inform} utterance types. 
In other words, \bagel{} is the most complex dataset to generate from. 
Even though it is more complex, metrics perform most reliably on \bagel{} here (note that the correlation is still only weak).
One possible explanation 
 is that \bagel{} only contains two human references per MR, whereas \sfh{} and \sfr{} both contain 5.35 references per MR on average. 
Having more references means that WBMs naturally will return higher scores (`anything goes'). This problem could possibly be solved by weighting multiple references according to their quality, as suggested by \cite{deltableu}, or following a reference-less approach \cite{specia:MT2010}.

 \noindent {\bf Quality of Data:}
 Our corpora contain crowdsourced human references that have grammatical errors, e.g.\ ``{\em Fifth Floor does not allow childs}'' (\sfr{} reference). 
 Corpus-based methods may pick up these errors, and word-based metrics will rate these system utterances as correct, whereas we can expect human judges to be sensitive to ungrammatical utterances. Note that the parsing score (while being a crude approximation of grammaticality) achieves one of our highest correlation results against human ratings, with  $|\rho|=.31$.
 Grammatical errors raise questions about the quality of the training data, especially when being crowdsourced.
 For example, \citet{Belz:EACL06} find that human experts assign low rankings to their original corpus text. Again,  weighting \cite{deltableu} or reference-less approaches \cite{specia:MT2010} might remedy this issue.
 
 \ignore{
 Following this approach, we asked crowdworkers to rate 
200 randomly sampled instances from each corpus, following the procedure outlined in Section \ref{sec:datacollection}. However, we find no significant correlation between rankings of the original references and the output generated by the systems. This suggests that data-driven approaches can cancel out noise. 
}

\begin{table*}[htp!]
\centering
\begin{adjustbox}{max width=1\textwidth}
\begin{tabular}{c|c|c|c}
& \multicolumn{2}{c|}{\bf Dimension of human ratings} & \\
{\bf Study} & {\bf Sentence Planning}  & {\bf Surface Realisation} & {\bf Domain} \\
\hline \hline
this paper  &  weak positive ($\rho=0.33$, WPS) & weak negative ($\rho=0.-31$, parser)& NLG, restaurant/hotel search \Tstrut\\
 
 \hline
\cite{belz:CL2009} &  none & strong positive (Pearson's $r=0.96$, \nist) & NLG, weather forecast \\
 
  \hline
\cite{stent2005evaluating}  &  weak positive ($\rho=0.47$, \textsc{\small LSA}) & negative ($\rho=-0.56$, \nist)& paraphrasing of news\\
   \hline
\cite{Liu:EMNLP2016}  & weak positive ($\rho=0.35$, \bleu{\small -4}) &  N/A & dialogue/Twitter pairs\\

\hline
\cite{keller:acl2014}  &  positive ($\rho=0.53$, \meteor) & N/A & image caption  \\

 \hline
\cite{erdem:eacl2017}   &  positive ($\rho=0.64$, \textsc{\small spice}) &  N/A & image caption\\

  \hline
\cite{cahill:acl2009} & N/A & negative ($\rho=-0.64$, \rouge) & NLG, German news texts \\

\hline
\cite{espinosa_further_2010} & weak positive ($\rho=0.43$, \ter) & positive ($\rho=0.62$, \bleu{\small -4}) & NLG, news texts \\

\end{tabular}
\end{adjustbox}
\vspace{-0.1cm}
\caption{Best correlation results achieved by our and previous work. 
Dimensions targeted towards  Sentence Planning include `accuracy', `adequacy', `correctness', `informativeness'. Dimensions for Surface Realisation include `clarity', `fluency', `naturalness'.
}\label{tab:previous}
\vspace{-0.2cm}
\end{table*}%

\subsection{Example-based Analysis}
\label{ssec:lingExmpl}

As shown in previous sections, word-based metrics moderately agree with humans on bad  quality output, but cannot distinguish output of good or medium quality. Table \ref{tab:examples} provides  examples from our three systems.\footnote{Please note that WBMs 
tend to match against the reference that is closest to the generated output. Therefore, we only include the closest match in Table \ref{tab:examples} for simplicity.} Again, we observe different behaviour between WOMs and \simil{} scores. 
In Example 1, \vlachos{} generates a grammatically correct English sentence, which represents the meaning of the MR well, and, as a result, this utterance received high human ratings (median = 6) for \textit{informativeness, naturalness} and \textit{quality}. 
However, WOMs rate this utterance low, i.e.\ scores of \bleu{}{\small 1-4}, \nist, \lepor, \cider, \rouge{} and \meteor{} normalised into the 1-6 range all stay below 1.5. This is because the system-generated utterance
has low overlap with the human/corpus references.  
Note that the \simil{} score is high (5), as it ignores human references and
computes distributional semantic similarity between the MR and the system output.
Examples~2 and~3 show outputs which 
receive low scores from both automatic metrics and humans. 
WOMs score these system outputs low due to little or no overlap  with human references, whereas humans are sensitive to ungrammatical output and missing information (the former is partially captured by GBMs). Examples~2 and~3 also illustrate inconsistencies in human ratings since system output~2 is clearly worse than {output~3 and both are rated by human with a median score of 1. 
Example~4 shows an output of the \wen{} system which is semantically very similar to the reference (\simil=4) and rated high by humans, but WOMs fail to capture this similarity.
GBMs show more accurate results for this utterance, with mean of readability scores 4 and parsing score 3.5.

\section{Related Work}\label{sec:related} 

Table \ref{tab:previous} summarises results published by previous studies in related fields which investigate the relation between human scores and automatic metrics. 
These studies mainly considered WBMs,  
while we are the first study to consider GBMs. Some studies ask users to provide separate ratings for surface realisation (e.g.\ asking about `clarity' or `fluency'), whereas other studies focus only on sentence planning  (e.g.\ `accuracy', `adequacy', or `correctness').
In general, correlations reported by previous work range from weak to strong. 
The results confirm that metrics can be reliable indicators at system-level \cite{belz:CL2009}, while they perform less reliably at sentence-level \cite{stent2005evaluating}. Also, the results show that the metrics capture realization better than sentence planning.
There is a general trend showing that best-performing metrics tend to be the more complex ones, combining word-overlap, semantic similarity and term frequency weighting. Note, however, that the majority of previous works do not report whether any of the metric correlations are significantly different from each other.

\section{Conclusions}\label{sec:discussion}

This paper shows that state-of-the-art automatic evaluation metrics for NLG systems do not sufficiently reflect human ratings, which stresses the need for human evaluations. This result is opposed to the current trend of relying on automatic evaluation identified in \cite{gkatzia:enlg2015}.

A detailed error analysis suggests that automatic metrics are particularly weak in distinguishing  outputs of medium and good quality, which can be partially attributed to the fact that human judgements and metrics are given on different scales.
We also show that metric performance is data- and system-specific.

Nevertheless, our results also suggest that 
automatic metrics can be useful 
for error analysis by helping to find cases where the system is performing poorly. In addition, we find reliable results on system-level, which suggests that metrics can be useful for system development.%

\section{Future Directions}\label{sec:future}

Word-based metrics make two strong assumptions: They treat human-generated references as a gold standard, which is \textit{correct} and \textit{complete}. 
We argue that these assumptions are invalid for corpus-based NLG, especially when using crowdsourced datasets. 
Grammar-based metrics, on the other hand, do not rely on human-generated references and are not influenced by their quality. However, these metrics can be easily manipulated with grammatically correct and easily readable output that is unrelated to the input. We have experimented with combining WBMs and GBMs using ensemble-based learning. However, while our model achieved high correlation with humans within a single domain, its cross-domain performance is insufficient.

Our paper clearly demonstrates the need for
more advanced metrics, as used in related fields, 
 including:  
assessing output quality within the dialogue context, e.g.\ \cite{duvsek-jurcicek:2016:SIGDIAL}; extrinsic evaluation metrics, such as NLG's contribution to task success, e.g.\ \cite{Rieser:IEEE14,Gkatzia:acl16,hastie:iwsds2016}; 
building discriminative models, e.g.\ \cite{hodosh2016focused}, \cite{vinyals:gans}; or reference-less quality prediction as used in MT, e.g.\ \cite{specia:MT2010}.
We see our paper as a first step towards reference-less evaluation for NLG by introducing grammar-based metrics. In current work
 \cite{dusek:icml2017}, we investigate a reference-less quality estimation approach based on recurrent neural networks, which predicts a quality score for a NLG system output by comparing it to the source meaning representation only.

Finally, note that the datasets considered in this study are fairly small (between  404 and 2.3k human references per domain). To remedy this, systems train on de-lexicalised versions \cite{wen:emnlp2015}, which bears the danger of ungrammatical lexicalisation \cite{SharmaHSSB16} and a possible overlap between testing and training set \cite{vlachos:coling2016}. There are ongoing efforts to release larger and more diverse data sets, e.g.\ \cite{novikova:INLG2016, novikova2017e2e}.

\section*{Acknowledgements}
This research received funding from the EPSRC projects  DILiGENt (EP/M005429/1) and  MaDrIgAL (EP/N017536/1). The Titan Xp used for this research was donated by the NVIDIA Corporation.

\bibliography{acl2017}
\bibliographystyle{emnlp_natbib}

\onecolumn
\section*{Appendix A: Detailed Results}

\bigskip
\begin{table*}[ht]
\centering
\begin{adjustbox}{max width=1\textwidth}
\begin{tabular}{c|l|c|c|c|l|c|l|c|c|c|l|c|l|c|c|c|l}
\multicolumn{6}{c|}{\textbf{\bagel}} & \multicolumn{6}{c|} {\textbf{\sfh}} & \multicolumn{6}{c}{\textbf{\sfr}} \\ 
\hline \hline
\multicolumn{2}{c|}{Inf: 0.16*} & \multicolumn{2}{c|}{Nat: 0.36*} & \multicolumn{2}{c|}{Qua: 0.38*} & \multicolumn{2}{c|}{Inf: 0.41*} & \multicolumn{2}{c|}{Nat: 0.47*} & \multicolumn{2}{c|}{Qua: 0.52*} & \multicolumn{2}{c|}{Inf: 0.35*} & \multicolumn{2}{c|}{Nat: 0.29*} & \multicolumn{2}{c}{Qua: 0.35*} \\ \hline
\multicolumn{3}{c|}{\dusek: 0.42*} & \multicolumn{3}{c|}{\vlachos: 0.24*} & \multicolumn{3}{c|}{\wen: 0.52*} & \multicolumn{3}{c|}{\vlachos: 0.45*}  & \multicolumn{3}{c|}{\wen: 0.28*}& \multicolumn{3}{c}{\vlachos: 0.38*} \\ 
\hline
\multicolumn{6}{c|}{Total \bagel: 0.31*} & \multicolumn{6}{c|}{Total \sfh: 0.50*} & \multicolumn{6}{c}{Total \sfr: 0.35*}\\ 
\hline
\multicolumn{18}{c}{Total all data: 0.45*}  \\
\end{tabular}
\end{adjustbox}

\medskip
\caption{Intra-class correlation coefficient (ICC) for human ratings across the three datasets. ``*'' denotes statistical significance ($p<0.05$).}
\label{stab:icc}
\end{table*}

\bigskip

\begin{table*}[hp]
\centering
\small
\begin{tabular}{l|ll|ll|ll}
 & \multicolumn{2}{c|}{\textbf{\bagel}} & \multicolumn{2}{c|}{\textbf{\sfh}} & \multicolumn{2}{c}{\textbf{\sfr}} \\
 & \dusek & \vlachos & \wen & \vlachos & \wen & \vlachos \\
\textit{metric} & \textit{Avg / StDev} & \textit{Avg / StDev} & \textit{Avg / StDev} & \textit{Avg / StDev} & \textit{Avg / StDev} & \textit{Avg / StDev} \\ \hline
\hline

\ter & 0.36/0.24&0.33/0.24&0.28*/0.27&0.65*/0.32&0.41*/0.35&0.65*/0.27 \\
\bleu1 & 0.75*/0.21&0.81*/0.16&0.85*/0.18&0.66*/0.23&0.73*/0.24&0.59*/0.23 \\
\bleu2 & 0.68/0.23&0.72/0.21&0.78*/0.25&0.54*/0.28&0.62*/0.31&0.45*/0.29 \\
\bleu3 & 0.60/0.28&0.63/0.26&0.69*/0.32&0.42*/0.33&0.52*/0.37&0.34*/0.33 \\
\bleu4 & 0.52/0.32&0.53/0.33&0.56*/0.40&0.28*/0.33&0.44*/0.41&0.24*/0.32 \\
\rouge & 0.76/0.18&0.78/0.17&0.83*/0.18&0.64*/0.21&0.72*/0.24&0.58*/0.22 
\\
\nist & 4.44*/2.05&4.91*/2.04&4.37*/2.19&3.49*/1.99&4.86*/2.55&4.01*/2.07 \\
\lepor & 0.46*/0.22&0.50*/0.19&0.52*/0.23&0.30*/0.16&0.51*/0.25&0.30*/0.17 \\
\cider & 2.92/2.40&3.01/2.27&3.08*/2.05&1.66*/1.67&3.39*/2.53&2.09*/1.73 \\
\meteor & 0.50/0.22&0.53/0.23&0.62*/0.27&0.44*/0.20&0.54*/0.28&0.41*/0.19 \\
\simil & 0.66/0.09&0.65/0.12&0.76*/0.15&0.73*/0.14&0.76/0.13&0.77/0.14 \\
\hdashline

\fres & 86.79/19.48&83.39/20.41&70.90/17.07&69.62/19.14&64.67/19.07&64.27/22.22 \\
msp & 0.04*/0.21&0.14*/0.37&0.68/0.78&0.69/0.77&0.78/0.82&0.85/0.89 \\
prs & 84.51*/25.78&93.30*/27.04&97.58*/32.58&107.90*/36.41&93.74/34.98&97.20/39.30\\
len & 38.20*/14.22&42.54*/14.11&49.06*/15.77&51.69*/17.30&53.27*/19.50&50.92*/18.74 \\
wps & 10.08*/3.10&10.94*/3.19&11.43*/3.63&12.07*/4.17&11.15*/4.37&10.52*/4.21 \\
sps & 13.15*/4.98&14.61*/5.13&16.03*/4.88&17.02*/5.90&16.39*/6.17&15.41*/5.92 \\
cpw & 3.77/0.60&3.88/0.59&4.34/0.58&4.36/0.63&4.86*/0.64&4.94*/0.76 \\
spw & 1.30/0.22&1.33/0.23&1.43/0.23&1.43/0.26&1.50/0.26&1.50/0.29 \\
pol & 2.22/1.21&2.40/1.16&1.24/1.04&1.33/1.04&1.69/1.12&1.57/1.07 \\
ppw & 0.22/0.09&0.22/0.09&0.11/0.10&0.12/0.09&0.16/0.11&0.16/0.12 \\
\hline
informativeness & 4.77/1.09&4.91/1.23&5.47*/0.81&5.27/1.02&5.29*/0.94&5.16/1.07 \\
naturalness & 4.76/1.26&4.67/1.25&4.99*/1.13&4.62/1.28&4.86/1.13&4.74/1.23 \\
quality & 4.77/1.19&4.54/1.28&4.54/1.18&4.53/1.26&4.51/1.14&4.58/1.33
\end{tabular}

\medskip
\caption{The systems' performance for all datasets. \textit{Avg} denotes a mean value, \textit{StDev} stands for standard deviation, ``*'' denotes a statistically significant difference ($p<0.05$) between the two systems on the given dataset.
}\label{stab:sys-eval}
\end{table*}

\ignore{
\begin{table*}[htp]
\centering
\begin{adjustbox}{max width=0.7\textwidth}
\begin{tabular}{l|ccc|c}
\multicolumn{1}{c}{} & \multicolumn{3}{|c|}{\vlachos} & \multicolumn{1}{c}{\wen} \\
 & \bagel-\sfh & \bagel-\sfr & \sfr-\sfh & \sfh-\sfr \\
 \hline \hline
\ter & 0.00 & 0.00 & 0.91 & 0.00 \\
\bleu1 & 0.00 & 0.00 & 0.00 & 0.00 \\
\bleu2 & 0.00 & 0.00 & 0.00 & 0.00 \\
\bleu3 & 0.00 & 0.00 & 0.00 & 0.00 \\
\bleu4 & 0.00 & 0.00 & 0.06 & 0.00 \\
\rouge & 0.00 & 0.00 & 0.00 & 0.00 \\
\nist & 0.00 & 0.00 & 0.00 & 0.00 \\
\lepor & 0.00 & 0.00 & 0.95 & 0.63 \\
\cider & 0.00 & 0.00 & 0.00 & 0.03 \\
\meteor & 0.00 & 0.00 & 0.00 & 0.00 \\
\simil & 0.00 & 0.00 & 0.00 & 0.61 \\
\hline
\fres & 0.00 & 0.00 & 0.00 & 0.00 \\
cpw & 0.00 & 0.00 & 0.00 & 0.00 \\
len & 0.00 & 0.00 & 0.51 & 0.00 \\
wps & 0.00 & 0.14 & 0.00 & 0.27 \\
sps & 0.00 & 0.07 & 0.00 & 0.29 \\
spw & 0.00 & 0.00 & 0.00 & 0.00 \\
pol & 0.00 & 0.00 & 0.00 & 0.00 \\
ppw & 0.00 & 0.00 & 0.00 & 0.00 \\
msp & 0.00 & 0.00 & 0.00 & 0.05 \\
prs & 0.00 & 0.12 & 0.00 & 0.06
\end{tabular}
\end{adjustbox}

\medskip
\caption{Significance levels ($p$-value) when comparing the performance of an individual system on different datasets. This illustrates how the results for each system differ depending on the target dataset.}
\label{stab:avg-sign}
\end{table*}
} 

\begin{sidewaystable}[thp]
\centering
\begin{adjustbox}{max width=1\textwidth}
\begin{tabular}{l|llllll|llllll|llllll}
\multicolumn{1}{c|}{} & \multicolumn{6}{c}{\bagel} & \multicolumn{6}{c|}{\sfh} & \multicolumn{6}{c}{\sfr} \\
\multicolumn{1}{c|}{} & \multicolumn{3}{c}{\dusek} & \multicolumn{3}{c|}{\vlachos} & \multicolumn{3}{c}{\wen} & \multicolumn{3}{c|}{\vlachos} & \multicolumn{3}{c}{\wen} & \multicolumn{3}{c}{\vlachos} \\
\textit{metric} & \textit{inf} & \textit{nat} & \textit{qual} & \textit{inf} & \textit{nat} & \textit{qual} & \textit{inf} & \textit{nat} & \textit{qual} & \textit{inf} & \textit{nat} & \textit{qual} & \textit{inf} & \textit{nat} & \textit{qual} & \textit{inf} & \textit{nat} & \textit{qual} \\
\hline \hline
\ter & -0.21* & -0.19* & -0.16* & -0.16* & -0.19* & -0.16* & -0.03 & -0.09 & -0.08 & -0.06 & \textbf{-0.20*} & -0.12* & 0.02 & -0.14* & -0.08 & \textbf{-0.16*} & -0.14* & -0.14* \\
\bleu1 & \textbf{0.30*} & 0.15* & 0.13 & 0.13 & 0.15* & 0.13 & 0.09 & 0.09* & 0.08 & 0.01 & 0.12* & 0.06 & 0.02 & 0.12* & 0.06 & \textbf{0.19*} & 0.15* & 0.13* \\
\bleu2 & \textbf{0.30*} & 0.17* & 0.14 & 0.12 & 0.14* & 0.11 & 0.08 & 0.09* & 0.07 & 0.00 & 0.12* & 0.07 & 0.01 & 0.13* & 0.07 & \textbf{0.14*} & 0.10* & 0.08* \\
\bleu3 & \textbf{0.27*} & 0.17* & 0.12 & 0.11 & 0.13 & 0.10 & 0.06 & 0.08 & 0.06 & 0.01 & 0.11* & 0.08 & 0.02 & 0.13* & 0.09* & 0.12* & 0.08 & 0.07 \\
\bleu4 & \textbf{0.23*} & 0.15* & 0.11 & 0.11 & 0.13 & 0.10 & 0.06 & 0.05 & 0.07 & 0.00 & 0.02 & 0.03 & 0.03 & 0.12* & 0.07 & 0.12* & 0.04 & 0.05 \\
\rouge & 0.20* & 0.11 & 0.09 & 0.20* & 0.17* & 0.15* & 0.07 & 0.09 & 0.08 & -0.01 & 0.04 & 0.02 & 0.04 & 0.17* & 0.09* & 0.12* & 0.11* & 0.08 \\
\nist & 0.24* & 0.07 & 0.02 & 0.16* & 0.13 & 0.11 & 0.07 & 0.05 & 0.01 & 0.02 & 0.14* & 0.11* & 0.03 & 0.07 & 0.01 & \textbf{0.15*} & 0.08 & 0.07 \\
\lepor & \textbf{0.17*} & 0.12 & \textbf{0.07} & -0.07 & 0.02 & -0.04 & 0.03 & 0.03 & 0.03 & 0.14* & \textbf{0.17*} & 0.10* & 0.00 & 0.05 & -0.02 & \textbf{0.28*} & 0.17* & 0.18* \\
\cider & \textbf{0.26*} & 0.14* & 0.10 & 0.14* & 0.19* & 0.14* & 0.07 & 0.07 & 0.00 & 0.03 & 0.13* & 0.09 & 0.02 & 0.12* & 0.03 & 0.10* & 0.11* & 0.08 \\
\meteor & 0.29* & 0.09 & 0.09 & 0.20* & 0.18* & 0.16* & 0.07 & 0.10* & 0.10* & 0.05 & 0.06 & 0.04 & 0.06 & 0.16* & 0.09* & \textbf{0.23*} & 0.19* & 0.17* \\
\simil & 0.16* & 0.04 & 0.06 & 0.14* & 0.13 & 0.09 & -0.05 & -0.12* & -0.11* & 0.03 & -0.03 & -0.08 & 0.13* & -0.06 & -0.08* & 0.19* & 0.01 & 0.02 \\
\hdashline
\fres & -0.06 & \textbf{0.09} & 0.13 & -0.09 & -0.04 & 0.04 & 0.00 & 0.03 & 0.10* & -0.01 & 0.03 & 0.09 & 0.00 & -0.05 & 0.02 & 0.04 & \textbf{0.09*} & 0.08* \\
cpw & 0.03 & \textbf{-0.12} & \textbf{-0.19*} & 0.08 & 0.05 & -0.03 & 0.02 & -0.02 & {\bf -0.09*} & \textbf{0.13*} & \textbf{0.14*} & 0.06 & 0.02 & 0.11* & 0.01 & 0.06 & 0.10* & 0.09* \\
len & \textbf{0.25*} & -0.25* & -0.21* & 0.04 & -0.19* & -0.24* & 0.01 & -0.17* & -0.09 & 0.12* & -0.08 & -0.07 & 0.11* & -0.17* & -0.08 & 0.21* & -0.14* & -0.09* \\
wps & \textbf{0.33*} & -0.17* & -0.12 & -0.05 & \textbf{-0.28*} & \textbf{-0.29*} & 0.01 & -0.15* & -0.05 & 0.08 & -0.12* & -0.08 & 0.11* & -0.19* & -0.07 & 0.18* & -0.15* & -0.11* \\
sps & \textbf{0.25*} & -0.20* & -0.17* & 0.03 & -0.17* & -0.23* & -0.02 & -0.16* & -0.08 & 0.02 & -0.18* & -0.16* & 0.07 & -0.17* & -0.08 & 0.12* & -0.21* & -0.16* \\
spw & 0.01 & -0.07 & \textbf{-0.13} & 0.10 & \textbf{0.09} & 0.02 & -0.08 & -0.02 & -0.11* & -0.10* & -0.10* & -0.17* & -0.07 & 0.06 & -0.03 & -0.14* & \textbf{-0.10*} & -0.11* \\
pol & 0.16* & -0.06 & -0.07 & 0.11 & -0.03 & -0.12 & -0.07 & -0.10* & -0.15* & 0.01 & -0.09 & -0.14* & -0.04 & -0.04 & -0.03 & -0.02 & -0.13* & -0.11* \\
ppw & -0.02 & 0.06 & 0.00 & \textbf{0.16*} & 0.15* & 0.08 & -0.09 & -0.06 & -0.16* & -0.02 & -0.01 & -0.09 & -0.09* & \textbf{0.08} & 0.00 & -0.13* & -0.05 & -0.07 \\
msp & -0.02 & -0.06 & -0.11 & 0.02 & -0.02 & -0.10 & -0.01 & -0.10* & -0.08 & 0.05 & -0.02 & -0.03 & 0.05 & 0.02 & -0.06 & 0.12* & 0.01 & 0.07 \\
prs & \textbf{-0.23*} & 0.18* & 0.13 & 0.05 & 0.24* & 0.31* & -0.02 & 0.13* & 0.09 & -0.13* & 0.05 & 0.04 & -0.11* & 0.15* & 0.11* & -0.16* & 0.20* & 0.16*
\end{tabular}
\end{adjustbox}
\caption{Spearman correlation between metrics and human ratings for individual datasets and systems. ``*'' denotes statistically significant correlation ($p<0.05$), bold font denotes significantly stronger correlation when comparing two systems on the same dataset.} 
\label{stab:corr}
\end{sidewaystable}

\ignore{
\begin{table*}[]
\centering
\begin{adjustbox}{max width=0.87\textwidth}
\begin{tabular}{l|lll|l|lll|l|lll|l}
\multicolumn{1}{c}{\textit{}} & \multicolumn{4}{c}{\textit{informativeness}} & \multicolumn{4}{c}{\textit{naturalness}} & \multicolumn{4}{c}{\textit{quality}} \\
\multicolumn{1}{c}{} & \multicolumn{3}{|c|}{\vlachos} & \multicolumn{1}{c|}{\wen} & \multicolumn{3}{c|}{\vlachos} & \multicolumn{1}{c|}{\wen} & \multicolumn{3}{c|}{\vlachos} & \multicolumn{1}{c}{\wen} \\
 & \rot{\bagel-\sfh} & \rot{\bagel-\sfr} & \rot{\sfr-\sfh} & \rot{\sfr-\sfh} & \rot{\bagel-\sfh} & \rot{\bagel-\sfr} & \rot{\sfr-\sfh} & \rot{\sfr-\sfh} & \rot{\bagel-\sfh} & \rot{\bagel-\sfr} & \rot{\sfr-\sfh} & \rot{\sfr-\sfh} \\
 \hline \hline
\ter & 0.06 & 1.00 & 0.06 & 0.43 & 0.91 & 0.40 & 0.34 & 0.35 & 0.43 & 0.68 & 0.71 & 0.97 \\
\bleu1 & 0.03 & 0.28 & \textbf{0.00} & 0.18 & 0.63 & 0.98 & 0.65 & 0.56 & 0.20 & 0.94 & 0.22 & 0.74 \\
\bleu2 & \textbf{0.04} & 0.70 & \textbf{0.01} & 0.27 & 0.72 & 0.43 & 0.66 & 0.50 & 0.44 & 0.61 & 0.79 & 0.91 \\
\bleu3 & 0.08 & 0.87 & 0.06 & 0.49 & 0.75 & 0.34 & 0.53 & 0.39 & 0.65 & 0.53 & 0.86 & 0.60 \\
\bleu4 & \textbf{0.04} & 0.95 & \textbf{0.03} & 0.51 & 0.06 & 0.12 & 0.75 & 0.24 & 0.22 & 0.44 & 0.64 & 0.98 \\
\rouge & \textbf{0.00} & 0.16 & \textbf{0.02} & 0.58 & \textbf{0.02} & 0.26 & 0.27 & 0.14 & \textbf{0.02} & 0.18 & 0.30 & 0.78 \\
\nist & \textbf{0.01} & 0.86 & \textbf{0.02} & 0.48 & 0.84 & 0.39 & 0.29 & 0.73 & 1.00 & 0.50 & 0.50 & 0.95 \\
\lepor & \textbf{0.00} & \textbf{0.00} & \textbf{0.01} & 0.59 & \textbf{0.01} & \textbf{0.01} & 0.97 & 0.74 & \textbf{0.01} & \textbf{0.00} & 0.18 & 0.38 \\
\cider & \textbf{0.05} & 0.48 & 0.19 & 0.43 & 0.31 & 0.16 & 0.70 & 0.32 & 0.33 & 0.28 & 0.93 & 0.70 \\
\meteor & \textbf{0.01} & 0.54 & \textbf{0.00} & 0.89 & \textbf{0.03} & 0.84 & \textbf{0.02} & 0.35 & \textbf{0.03} & 0.86 & \textbf{0.02} & 0.82 \\
\simil & 0.07 & 0.41 & \textbf{0.01} & \textbf{0.00} & \textbf{0.00} & \textbf{0.04} & 0.43 & 0.30 & \textbf{0.00} & 0.21 & 0.10 & 0.57 \\
\hline
\fres & 0.16 & \textbf{0.02} & 0.38 & 0.96 & 0.26 & \textbf{0.02} & 0.26 & 0.15 & 0.39 & 0.45 & 0.92 & 0.17 \\
cpw & 0.34 & 0.78 & 0.22 & 0.95 & 0.14 & 0.45 & 0.46 & \textbf{0.02} & 0.14 & \textbf{0.04} & 0.59 & 0.07 \\
len & 0.14 & \textbf{0.00} & 0.09 & 0.07 & \textbf{0.05} & 0.37 & 0.29 & 0.99 & \textbf{0.00} & \textbf{0.01} & 0.67 & 0.84 \\
wps & \textbf{0.03} & \textbf{0.00} & 0.07 & 0.09 & \textbf{0.00} & \textbf{0.02} & 0.52 & 0.47 & \textbf{0.00} & \textbf{0.00} & 0.56 & 0.73 \\
sps & 0.77 & 0.12 & 0.07 & 0.10 & 0.89 & 0.57 & 0.66 & 0.90 & 0.25 & 0.25 & 1.00 & 0.97 \\
spw & \textbf{0.00} & \textbf{0.00} & 0.52 & 0.85 & \textbf{0.00} & \textbf{0.00} & 0.94 & 0.15 & \textbf{0.00} & \textbf{0.03} & 0.30 & 0.15 \\
pol & 0.10 & \textbf{0.03} & 0.56 & 0.60 & 0.30 & 0.07 & 0.42 & 0.26 & 0.71 & 0.93 & 0.64 & \textbf{0.04} \\
ppw & \textbf{0.00} & \textbf{0.00} & 0.06 & 0.95 & \textbf{0.00} & \textbf{0.00} & 0.48 & \textbf{0.02} & \textbf{0.00} & \textbf{0.01} & 0.75 & \textbf{0.01} \\
msp & 0.64 & 0.10 & 0.24 & 0.31 & 0.94 & 0.55 & 0.60 & \textbf{0.04} & 0.26 & \textbf{0.00} & 0.06 & 0.69 \\
prs & \textbf{0.00} & \textbf{0.00} & 0.59 & 0.10 & \textbf{0.00} & 0.46 & \textbf{0.01} & 0.65 & \textbf{0.00} & \textbf{0.01} & \textbf{0.04} & 0.68
\end{tabular}
\end{adjustbox}

\medskip
\caption{$p$-values of Williams test comparing Spearman correlations between datasets for individual systems (as presented in Table~\ref{stab:corr}). Bold font denotes significant difference ($p<0.05$).
}
\label{stab:corr-sign}
\end{table*}
} 

\begin{table*}[]
\centering
\begin{adjustbox}{max width=0.75\textwidth}
\begin{tabular}{l|lll|lll|lll}
 & \multicolumn{3}{c|}{\textbf{\bagel}} & \multicolumn{3}{c|}{\textbf{\sfh}} & \multicolumn{3}{c}{\textbf{\sfr}} \\
 & \multicolumn{1}{|c}{\textit{inf}} & \multicolumn{1}{c}{\textit{nat}} & \multicolumn{1}{c|}{\textit{qual}} & \multicolumn{1}{c}{\textit{inf}} & \multicolumn{1}{c}{\textit{nat}} & \multicolumn{1}{c|}{\textit{qual}} & \multicolumn{1}{c}{\textit{inf}} & \multicolumn{1}{c}{\textit{nat}} & \multicolumn{1}{c}{\textit{qual}} \\
 \hline \hline
\ter & -0.19* & -0.19* & -0.15* & -0.10* & -0.19* & -0.07* & -0.09* & -0.15* & -0.08* \\
\bleu1 & 0.23* & 0.14* & 0.11* & 0.11* & 0.18* & 0.07* & 0.11* & 0.14* & 0.07* \\
\bleu2 & 0.21* & 0.15* & 0.12* & 0.10* & 0.17* & 0.07* & 0.09* & 0.13* & 0.06* \\
\bleu3 & 0.19* & 0.15* & 0.11* & 0.09* & 0.16* & 0.07* & 0.08* & 0.12* & 0.06* \\
\bleu4 & 0.18* & 0.14* & 0.10* & 0.08* & 0.10* & 0.06 & 0.09* & 0.09* & 0.05 \\
\rouge & 0.20* & 0.13* & 0.11* & 0.09* & 0.15* & 0.06 & 0.09* & 0.15* & 0.06* \\
\nist & 0.21* & 0.09 & 0.06 & 0.07* & 0.13* & 0.06 & 0.10* & 0.08* & 0.03 \\
\lepor & 0.07 & 0.07 & 0.01 & 0.13* & 0.15* & 0.05 & 0.16* & 0.12* & 0.04 \\
\cider & 0.21* & 0.16* & 0.12* & 0.10* & 0.16* & 0.05 & 0.08* & 0.12* & 0.04 \\
\meteor & 0.25* & 0.13* & 0.12* & 0.11* & 0.15* & 0.08* & 0.15* & 0.18* & 0.11* \\
\simil & 0.15* & 0.09 & 0.07 & 0.01 & -0.04 & -0.09* & 0.15* & -0.02 & -0.02 \\
\hline
\fres & -0.08 & 0.03 & 0.09 & 0.01 & 0.04 & 0.10* & 0.02 & 0.02 & 0.06 \\
cpw & 0.05 & -0.04 & -0.12* & 0.07* & 0.05 & -0.02 & 0.04 & 0.10* & 0.06 \\
len & 0.14* & -0.22* & -0.24* & 0.05 & -0.14* & -0.07* & 0.16* & -0.15* & -0.09* \\
wps & 0.14* & -0.23* & -0.23* & 0.03 & -0.14* & -0.06 & 0.14* & -0.17* & -0.10* \\
sps & 0.14* & -0.19* & -0.21* & -0.01 & -0.18* & -0.12* & 0.10* & -0.18* & -0.12* \\
spw & 0.05 & 0.00 & -0.06 & -0.10* & -0.06 & -0.14* & -0.11* & -0.02 & -0.07* \\
pol & 0.13* & -0.05 & -0.10* & -0.04 & -0.10* & -0.14* & -0.03 & -0.08* & -0.08* \\
ppw & 0.06 & 0.11* & 0.04 & -0.06 & -0.04 & -0.13* & -0.11* & 0.01 & -0.04 \\
msp & 0.02 & -0.04 & -0.11* & 0.02 & -0.06 & -0.06 & 0.08* & 0.01 & 0.01 \\
prs & -0.10 & 0.22* & 0.25* & -0.05 & 0.12* & 0.07 & -0.13* & 0.18* & 0.13*
\end{tabular}
\end{adjustbox}

\medskip
\caption{Spearman correlation between metrics and human ratings for each dataset. ``*'' denotes statistically significant correlation ($p<0.05$).}
\label{stab:corr-ds}
\end{table*}

\ignore{
\begin{table*}[]
\centering
\begin{adjustbox}{max width=0.68\textwidth}
\begin{tabular}{l|lll|lll|lll}
\multicolumn{1}{c}{} & \multicolumn{3}{c}{informativeness} & \multicolumn{3}{c}{naturalness} & \multicolumn{3}{c}{quality} \\
\multicolumn{1}{c|}{} & \multicolumn{1}{c}{\rot{\bagel-\sfh}} & \multicolumn{1}{c}\rot{\rot{\bagel-\sfr}} & \multicolumn{1}{c|}{\rot{\sfr-\sfh}} & \multicolumn{1}{c}{\rot{\bagel-\sfh}} & \multicolumn{1}{c}{\rot{\bagel-\sfr}} & \multicolumn{1}{c|}{\rot{\sfr-\sfh}} & \multicolumn{1}{c}{\rot{\bagel-\sfh}} & \multicolumn{1}{c}{\rot{\bagel-\sfr}} & \multicolumn{1}{c}{\rot{\sfh-\sfr}} \\
\hline \hline
\ter & 0.12 & 0.07 & 0.81 & 0.90 & 0.48 & 0.41 & 0.16 & 0.19 & 0.93 \\
\bleu1 & \textbf{0.03} & \textbf{0.04} & 0.96 & 0.54 & 0.99 & 0.55 & 0.44 & 0.44 & 1.00 \\
\bleu2 & \textbf{0.04} & \textbf{0.03} & 0.87 & 0.69 & 0.65 & 0.39 & 0.44 & 0.32 & 0.83 \\
\bleu3 & 0.07 & 0.05 & 0.91 & 0.84 & 0.55 & 0.42 & 0.49 & 0.40 & 0.89 \\
\bleu4 & 0.11 & 0.11 & 0.99 & 0.52 & 0.41 & 0.85 & 0.43 & 0.36 & 0.89 \\
\rouge & 0.06 & 0.05 & 0.98 & 0.81 & 0.74 & 0.93 & 0.39 & 0.39 & 0.99 \\
\nist & \textbf{0.02} & 0.06 & 0.60 & 0.53 & 0.90 & 0.45 & 0.93 & 0.69 & 0.62 \\
\lepor & 0.29 & 0.12 & 0.61 & 0.12 & 0.36 & 0.54 & 0.47 & 0.56 & 0.90 \\
\cider & 0.05 & \textbf{0.02} & 0.78 & 0.91 & 0.49 & 0.56 & 0.24 & 0.14 & 0.77 \\
\meteor & \textbf{0.01} & 0.07 & 0.46 & 0.72 & 0.35 & 0.57 & 0.55 & 0.94 & 0.60 \\
\simil & \textbf{0.01} & 0.95 & \textbf{0.01} & \textbf{0.03} & 0.05 & 0.82 & \textbf{0.00} & 0.12 & 0.20 \\
\hline
\fres & 0.13 & 0.08 & 0.82 & 0.93 & 0.78 & 0.71 & 0.88 & 0.54 & 0.44 \\
cpw & 0.74 & 0.86 & 0.61 & 0.09 & 0.01 & 0.37 & 0.09 & \textbf{0.00} & 0.19 \\
len & 0.12 & 0.66 & \textbf{0.04} & 0.15 & 0.23 & 0.82 & \textbf{0.00} & \textbf{0.00} & 0.84 \\
wps & 0.07 & 0.92 & 0.05 & 0.11 & 0.27 & 0.62 & \textbf{0.00} & \textbf{0.02} & 0.50 \\
sps & \textbf{0.01} & 0.49 & 0.05 & 0.92 & 0.94 & 0.98 & 0.08 & 0.10 & 0.88 \\
spw & \textbf{0.01} & \textbf{0.01} & 0.87 & 0.23 & 0.69 & 0.43 & 0.14 & 0.79 & 0.23 \\
pol & \textbf{0.00} & \textbf{0.01} & 0.84 & 0.36 & 0.57 & 0.73 & 0.46 & 0.65 & 0.23 \\
ppw & \textbf{0.03} & \textbf{0.00} & 0.40 & 0.01 & 0.11 & 0.35 & \textbf{0.00} & 0.18 & 0.12 \\
msp & 0.97 & 0.29 & 0.28 & 0.76 & 0.33 & 0.20 & 0.37 & \textbf{0.03} & 0.19 \\
prs & 0.37 & 0.52 & 0.13 & 0.08 & 0.51 & 0.28 & \textbf{0.00} & \textbf{0.03} & 0.27
\end{tabular}
\end{adjustbox}

\medskip
\caption{$p$-values of Williams test comparing Spearman correlations between datasets (as presented in Table~\ref{stab:corr-ds}). Bold font denotes significance ($p<0.05$).}
\label{stab:corr-ds-sign}
\end{table*}
} 

\begin{table*}[ht]
\centering
\begin{adjustbox}{max width=0.75\textwidth}
\begin{tabular}{l|lll|lll|lll}
 & \multicolumn{3}{c|}{\textbf{\dusek}} & \multicolumn{3}{c|}{\textbf{\vlachos}} & \multicolumn{3}{c}{\textbf{\wen}} \\
 & \multicolumn{1}{|c}{\textit{inf}} & \multicolumn{1}{c}{\textit{nat}} & \multicolumn{1}{c|}{\textit{qual}} & \multicolumn{1}{c}{\textit{inf}} & \multicolumn{1}{c}{\textit{nat}} & \multicolumn{1}{c|}{\textit{qual}} & \multicolumn{1}{c}{\textit{inf}} & \multicolumn{1}{c}{\textit{nat}} & \multicolumn{1}{c}{\textit{qual}} \\
 \hline \hline
\ter & -0.21* & -0.19* & -0.16* & -0.07* & -0.15* & -0.11* & -0.02 & -0.13* & -0.08* \\
\bleu1 & 0.30* & 0.15* & 0.13 & 0.08* & 0.12* & 0.08* & 0.07* & 0.13* & 0.07* \\
\bleu2 & 0.30* & 0.17* & 0.14 & 0.05 & 0.11* & 0.07* & 0.06* & 0.14* & 0.08* \\
\bleu3 & 0.27* & 0.17* & 0.12 & 0.04 & 0.09* & 0.07* & 0.06 & 0.13* & 0.08* \\
\bleu4 & 0.23* & 0.15* & 0.11 & 0.04 & 0.04 & 0.04 & 0.06 & 0.11* & 0.08* \\
\rouge & 0.20* & 0.11 & 0.09 & 0.05 & 0.09* & 0.05 & 0.07* & 0.15* & 0.09* \\
\nist & 0.25* & 0.07 & 0.02 & 0.07* & 0.11* & 0.09* & 0.04 & 0.06* & 0.01 \\
\lepor & 0.17* & 0.12 & 0.07 & 0.13* & 0.13* & 0.11* & 0.02 & 0.05 & 0.00 \\
\cider & 0.26* & 0.14* & 0.10 & 0.05 & 0.13* & 0.09* & 0.04 & 0.10* & 0.02 \\
\meteor & 0.29* & 0.09 & 0.09 & 0.14* & 0.13* & 0.12* & 0.08* & 0.15* & 0.10* \\
\simil & 0.16* & 0.04 & 0.06 & 0.14* & 0.02 & 0.00 & 0.05 & -0.08* & -0.09* \\
\hline \hline
\fres & -0.06 & 0.09 & 0.13 & -0.02 & 0.04 & 0.07* & 0.02 & -0.01 & 0.06* \\
cpw & 0.03 & -0.12 & -0.19* & 0.11* & 0.11* & 0.08* & -0.02 & 0.02 & -0.05 \\
len & 0.25* & -0.25* & -0.21* & 0.17* & -0.12* & -0.10* & 0.06 & -0.18* & -0.08* \\
wps & 0.33* & -0.17* & -0.12 & 0.11* & -0.17* & -0.13* & 0.07* & -0.17* & -0.06 \\
sps & 0.25* & -0.20* & -0.17* & 0.09* & -0.19* & -0.17* & 0.03 & -0.17* & -0.08* \\
spw & 0.01 & -0.07 & -0.13 & -0.07* & -0.06* & -0.10* & -0.09* & 0.01 & -0.07* \\
pol & 0.16* & -0.06 & -0.07 & -0.02 & -0.09* & -0.11* & -0.08* & -0.08* & -0.09* \\
ppw & -0.02 & 0.06 & 0.00 & -0.08* & 0.00 & -0.05 & -0.11* & 0.00 & -0.07* \\
msp & -0.02 & -0.06 & -0.11 & 0.10* & 0.00 & 0.02 & 0.02 & -0.04 & -0.07* \\
prs & -0.23* & 0.18* & 0.13 & -0.12* & 0.16* & 0.15* & -0.07* & 0.14* & 0.10*
\end{tabular}
\end{adjustbox}

\medskip
\caption{Spearman correlation between metrics and human ratings for each system. ``*'' denotes statistical significance ($p<0.05$).}
\label{stab:corr-sys}
\end{table*}

\ignore{
\begin{table*}[h]
\centering
\begin{adjustbox}{max width=0.68\textwidth}
\begin{tabular}{l|lll|lll|lll}
\multicolumn{1}{c}{} & \multicolumn{3}{c}{informativeness} & \multicolumn{3}{c}{naturalness} & \multicolumn{3}{c}{quality} \\
 & \rot{\dusek-\vlachos} & \rot{\dusek-\wen} & \rot{\vlachos-\wen} & \rot{\dusek-\vlachos} & \rot{\dusek-\wen} & \rot{\vlachos-\wen} & \rot{\dusek-\vlachos} & \rot{\dusek-\wen} & \rot{\vlachos-\wen} \\
 \hline \hline
\ter & \textbf{0.01} & \textbf{0.00} & 0.42 & 0.45 & 0.29 & 0.76 & 0.43 & 0.18 & 0.58 \\
\bleu1 & \textbf{0.00} & \textbf{0.00} & 0.88 & 0.61 & 0.74 & 0.87 & 0.45 & 0.35 & 0.86 \\
\bleu2 & \textbf{0.00} & \textbf{0.00} & 0.84 & 0.25 & 0.55 & 0.59 & 0.27 & 0.31 & 0.92 \\
\bleu3 & \textbf{0.00} & \textbf{0.00} & 0.84 & 0.17 & 0.46 & 0.53 & 0.35 & 0.46 & 0.85 \\
\bleu4 & \textbf{0.00} & \textbf{0.00} & 0.72 & \textbf{0.05} & 0.40 & 0.24 & 0.27 & 0.60 & 0.56 \\
\rouge & \textbf{0.01} & \textbf{0.02} & 0.74 & 0.76 & 0.41 & 0.26 & 0.58 & 0.87 & 0.47 \\
\nist & \textbf{0.00} & \textbf{0.00} & 0.60 & 0.47 & 0.92 & 0.41 & 0.21 & 0.79 & 0.13 \\
\lepor & 0.43 & \textbf{0.01} & 0.05 & 0.78 & 0.23 & 0.14 & 0.53 & 0.22 & 0.06 \\
\cider & \textbf{0.00} & \textbf{0.00} & 0.78 & 0.89 & 0.47 & 0.55 & 0.85 & 0.12 & 0.18 \\
\meteor & \textbf{0.00} & \textbf{0.00} & 0.31 & 0.48 & 0.31 & 0.77 & 0.64 & 0.86 & 0.77 \\
\simil & 0.85 & 0.07 & 0.10 & 0.73 & \textbf{0.03} & 0.06 & 0.35 & \textbf{0.01} & 0.10 \\
\hline
\fres & 0.46 & 0.15 & 0.49 & 0.36 & 0.08 & 0.40 & 0.29 & 0.22 & 0.87 \\
cpw & 0.17 & 0.32 & \textbf{0.02} & \textbf{0.00} & \textbf{0.02} & 0.09 & \textbf{0.00} & \textbf{0.01} & \textbf{0.03} \\
len & 0.16 & \textbf{0.00} & \textbf{0.05} & \textbf{0.02} & 0.21 & 0.31 & 0.05 & \textbf{0.02} & 0.74 \\
wps & \textbf{0.00} & \textbf{0.00} & 0.46 & 0.92 & 0.99 & 0.91 & 0.78 & 0.30 & 0.19 \\
sps & \textbf{0.00} & \textbf{0.00} & 0.31 & 0.90 & 0.60 & 0.69 & 0.96 & 0.11 & 0.10 \\
spw & 0.21 & 0.09 & 0.65 & 0.85 & 0.13 & 0.18 & 0.64 & 0.28 & 0.54 \\
pol & \textbf{0.00} & \textbf{0.00} & 0.30 & 0.62 & 0.77 & 0.84 & 0.47 & 0.80 & 0.64 \\
ppw & 0.31 & 0.10 & 0.52 & 0.31 & 0.31 & 0.99 & 0.48 & 0.23 & 0.63 \\
msp & \textbf{0.03} & 0.51 & 0.14 & 0.24 & 0.63 & 0.49 & \textbf{0.02} & 0.52 & 0.09 \\
prs & \textbf{0.04} & \textbf{0.01} & 0.45 & 0.73 & 0.44 & 0.67 & 0.73 & 0.59 & 0.37
\end{tabular}
\end{adjustbox}

\medskip
\caption{$p$-values of Williams test comparing Spearman correlations between systems (as presented in Table~\ref{stab:corr-sys}). Bold font denotes significance ($p<0.05$).}
\label{stab:corr-sys-sign}
\end{table*}
} 

\begin{sidewaystable}[htp]
\centering
\begin{adjustbox}{max width=1\textwidth}
\begin{tabular}{l|l|llllllllllll}
Accuracy       & rand  & \ter   & \bleu 1 & \bleu 2 & \bleu 3 & \bleu 4 & \rouge & \nist  & \lepor & \cider & {\small\textsc{mete}} & \fres  & \simil   \\
\hline
\bagel       &  &    &  &  &  &  &  &   &  & &  &   &    \\
inform & 37.13 & 45.05* & 41.58*   & 41.58*   & 42.57*   & 42.08*   & 43.07*    & 43.07* & 41.58* & 43.07* & 45.54*  & 37.13 & 41.09* \\
natural     & 42.08 & 47.03* & 46.04*   & 45.54*   & 44.06*   & 45.05*   & 46.04*    & 44.55* & 46.53* & 45.05* & 45.05*  & 42.08 & 43.07* \\
quality         & 33.17 & 45.54* & 43.07*   & 40.10*    & 40.59*   & 43.56*   & 43.07*    & 41.09* & 40.59* & 42.08* & 41.58*  & 37.62 & 42.57* \\

\hline
\sfh       &  &    &  &  &  &  &  &   &  & &  &   &    \\
inform & 25.38 & 34.92* & 35.68* & 35.18* & 35.68* & 34.67* & 36.43* & 31.41 & 32.16* & 33.92* & 36.43* & 34.92* & 33.92* \\
natural     & 41.96 & 45.73 & 46.48 & 45.48 & 46.48 & 45.23 & 48.74 & 41.21 & 43.72 & 44.72 & 49.75* & 37.19 & 46.98 \\
quality         & 44.47 & 40.95 & 40.95 & 42.21 & 44.72 & 41.46 & 43.22 & 40.2  & 40.95 & 42.46 & 45.98 & 33.67 & 37.44 \\

\hline
\sfr       &  &    &  &  &  &  &  &   &  & &  &   &    \\
inform & 33.68 & 36.27 & 35.41 & 34.02 & 34.72 & 36.96 & 33.16 & 35.58 & 36.27 & 32.47 & 34.72 & 38.34* & 42.66* \\
natural     & 36.10  & 40.41 & 40.07 & 38.86 & 38.34 & 38.86 & 38.17 & 39.38 & 41.11* & 36.79 & 39.38 & 39.38 & 38.00 \\
quality         & 39.38 & 37.13 & 36.96 & 39.21 & 37.65 & 39.55 & 36.10  & 38.69 & 39.72 & 35.23 & 34.89 & 40.93 & 37.31 \\

\hline
\sfr, quant.       &   &    &  &  &  &  &  &   &  & &  &   &    \\
inform & 31.95 & 35.75* & 36.27* & 34.37* & 35.92* & 34.54* & 36.44* & 39.55* & 37.13* & 36.27* & 36.79* & 38.17* & 42.83* \\
quality         & 39.21 & 33.33 & 34.37 & 32.3  & 30.57 & 26.94 & 34.54 & 33.16 & 35.92 & 30.92 & 31.61 & 32.47 & 35.41 \\
naturalness     & 37.13 & 37.82 & 38.69 & 36.1  & 35.75 & 32.3  & 36.96 & 39.21 & 38.86 & 35.23 & 38.34 & 34.2  & 36.1 
\end{tabular}
\end{adjustbox}

\medskip
\caption{Accuracy of metrics predicting relative human ratings, with ``*'' denoting statistical significance ($p<0.05$).}
 \label{tab:accuracy0}
\end{sidewaystable}

\begin{table*}[ht]
\centering
\begin{tabular}{l|lc|lc|lc}
 & \multicolumn{2}{c|}{\textit{informativeness}} & \multicolumn{2}{c|}{\textit{naturalness}} & \multicolumn{2}{c}{\textit{quality}} \\
 & \textit{Bad} & \textit{Good and avg} & \textit{Bad} & \textit{Good and avg} & \textit{Bad} & \textit{Good and avg} \\
 \hline \hline
\ter & \textbf{0.48*} & 0.07* & \textbf{0.31*} & 0.15* & 0.08 & 0.06* \\
\bleu1 & \textbf{0.45*} & 0.11* & \textbf{0.26*} & 0.13* & 0.07 & 0.04 \\
\bleu2 & \textbf{0.49*} & 0.09* & \textbf{0.29*} & 0.13* & 0.05 & 0.04* \\
\bleu3 & \textbf{0.40*} & 0.08* & \textbf{0.25*} & 0.13* & 0.01 & 0.05* \\
\bleu4 & \textbf{0.41*} & 0.07* & \textbf{0.21*} & 0.08* & 0.01 & 0.04 \\
\rouge & \textbf{0.50*} & 0.08* & \textbf{0.28*} & 0.13* & 0.07 & 0.04* \\
\nist & \textbf{0.26} & 0.08* & \textbf{0.23*} & 0.08* & 0.08 & 0.03 \\
\lepor & \textbf{0.40*} & 0.09* & \textbf{0.23*} & 0.10* & 0.03 & 0.01 \\
\cider & \textbf{0.42*} & 0.09* & \textbf{0.21*} & 0.12* & 0.02 & 0.04 \\
\meteor & \textbf{0.45*} & 0.14* & 0.24* & 0.15* & 0.03 & 0.08* \\
\simil & \textbf{0.37*} & 0.12* & \textbf{0.29*} & -0.03 & \textbf{0.21*} & -0.08*
\end{tabular}
\caption{Spearman correlation between WBM scores and human ratings for utterances from the \emph{Bad} bin and utterances from the \emph{Good} and \emph{Average} bins. ``*'' denotes statistically significant correlation ($p<0.05$), bold font denotes significantly stronger correlation for the \emph{Bad} bin compared to the \emph{Good} and \emph{Average} bins.}
\label{stab:corr-bins}
\end{table*}

\begin{table*}[]
\centering
\begin{tabular}{l|lc|lc|lc}
\multicolumn{1}{c}{\textit{}} & \multicolumn{2}{|c|}{\textit{informativeness}} & \multicolumn{2}{c}{\textit{naturalness}} & \multicolumn{2}{|c}{\textit{quality}} \\
\textit{} & \textit{Inform} & \textit{Not inform} & \textit{Inform} & \textit{Not inform} & \textit{Inform} & \textit{Not inform} \\
\hline \hline
\ter & -0.08* & -0.10 & -0.17* & -0.18* & -0.09* & -0.11* \\
\bleu1 & 0.11* & 0.09 & 0.14* & 0.20* & 0.07* & 0.11* \\
\bleu2 & 0.09* & 0.10 & 0.14* & 0.20* & 0.07* & 0.13* \\
\bleu3 & 0.07* & 0.11* & 0.13* & 0.20* & 0.06* & 0.14* \\
\bleu4 & 0.06* & 0.11* & 0.09* & 0.18* & 0.05* & 0.14* \\
\rouge & 0.08* & 0.12* & 0.14* & 0.22* & 0.06* & 0.16* \\
\nist & 0.08* & 0.05 & 0.10* & 0.06 & 0.07* & -0.06 \\
\lepor & 0.09* & 0.16* & 0.11* & 0.16* & 0.05* & 0.04 \\
\cider & 0.10* & 0.01 & \textbf{0.16*} & 0.04 & \textbf{0.07*} & 0.02 \\
\meteor & 0.14* & 0.17* & 0.15* & 0.22* & 0.09* & 0.18* \\
\simil & 0.15* & 0.09 & -0.01 & -0.03 & -0.05* & -0.10 \\
\hline
cpw & \textbf{0.12*} & -0.15* & \textbf{0.09*} & -0.14* & \textbf{0.01} & -0.11* \\
len & 0.17* & 0.08 & -0.15* & -0.12* & -0.12* & -0.05 \\
wps & 0.11* & 0.19* & \textbf{-0.19*} & -0.03 & \textbf{-0.12*} & 0.01 \\
sps & 0.09* & 0.18* & \textbf{-0.20*} & -0.02 & \textbf{-0.17*} & 0.02 \\
spw & -0.06* & 0.09 & -0.03 & 0.01 & \textbf{-0.12*} & 0.01 \\
pol & \textbf{-0.08*} & 0.05 & -0.10* & -0.03 & -0.09* & -0.03 \\
ppw & \textbf{-0.14*} & -0.01 & 0.00 & -0.03 & -0.03 & -0.05 \\
msp & \textbf{0.11*} & -0.03 & 0.00 & -0.08 & -0.03 & -0.08 \\
prs & -0.10* & -0.18* & \textbf{0.18*} & 0.04 & \textbf{0.15*} & 0.02
\end{tabular}
\caption{Spearman correlation between automatic metrics and human ratings for utterances of the \emph{inform} MR type and utterances of other MR types. ``*'' denotes statistically significant correlation ($p<0.05$), bold font denotes significantly stronger (absolute) correlation for \emph{inform} MRs compared to non-\emph{inform} MRs.}
\label{stab:corr-mr}
\end{table*}

\end{document}